\documentclass{article}

\usepackage{amsmath}
\usepackage{microtype}
\usepackage{graphicx}
\usepackage{subfigure}
\usepackage{booktabs} % for professional tables
\usepackage{multirow}
\usepackage{xcolor}
\usepackage{tikz}
\usepackage{pifont}
\def\checkmark{\tikz\fill[scale=0.4](0,.35) -- (.25,0) -- (1,.7) -- (.25,.15) -- cycle;} 
\usepackage{amsfonts}

\usepackage{mathtools}
\DeclareMathOperator*{\argmaxA}{arg\,max}

\usepackage{hyperref}

\usepackage[accepted]{sysml2019}

\sysmltitlerunning{A View on Deep Reinforcement Learning in System Optimization}

\begin{document}

\twocolumn[
\sysmltitle{A View on Deep Reinforcement Learning in System Optimization}

% unless you've provided the [accepted] option to the sysml2019


%\sysmlsetsymbol{equal}{*}

\begin{sysmlauthorlist}
\sysmlauthor{Ameer Haj-Ali}{intel,berkeley}
\sysmlauthor{Nesreen K. Ahmed}{intel}
\sysmlauthor{Ted Willke}{intel}
\sysmlauthor{Joseph E. Gonzalez}{berkeley}
\sysmlauthor{Krste Asanovic}{berkeley}
\sysmlauthor{Ion Stoica}{berkeley}
\end{sysmlauthorlist}

\sysmlaffiliation{intel}{Intel Labs}
\sysmlaffiliation{berkeley}{University of California, Berkeley}

\sysmlcorrespondingauthor{Ameer Haj-Ali}{ameerh@berkeley.edu}
%\sysmlcorrespondingauthor{Eee Pppp}{ep@eden.co.uk}

\sysmlkeywords{Machine Learning, SysML}

\vskip 0.3in

\vspace{-0.5cm}

\begin{abstract}
Many real-world systems problems require reasoning about the long term consequences of actions taken to configure and manage the system.  These problems with delayed and often sequentially aggregated reward, are often inherently reinforcement learning problems and present the opportunity to leverage the recent substantial advances in deep reinforcement learning. 
However, in some cases, it is not clear why deep reinforcement learning is a good fit for the problem. Sometimes, it does not perform better than the state-of-the-art solutions. And in other cases, random search or greedy algorithms could outperform deep reinforcement learning. In this paper, we review, discuss, and evaluate the recent trends of using deep reinforcement learning in system optimization. We propose a set of essential metrics to guide future works in evaluating the efficacy of using deep reinforcement learning in system optimization. Our evaluation includes challenges, the types of problems, their formulation in the deep reinforcement learning setting, embedding, the model used, efficiency, and robustness. We conclude with a discussion on open challenges and potential directions for pushing further the integration of reinforcement learning in system optimization.
\vspace{-0.4cm}

\end{abstract}
]
\printAffiliationsAndNotice{} 
\setlength\tabcolsep{0pt}
 \section{Introduction}
Reinforcement learning (RL) is a class of learning problems framed in the context of planning on a Markov Decision Process (MDP)~\cite{bellman1957}, when the MDP is not known. In RL, an agent continually interacts with the environment~\cite{kaelbling1996reinforcement,sutton2018reinforcement}. In particular, the agent observes the state of the environment, and based on this observation takes an action. The goal of the RL agent is then to compute a policy--a mapping between the environment states and actions--that maximizes a long term reward. There are multiple ways to extrapolate the policy. Non-approximation methods usually fail to predict good actions in states that were not visited in the past, and require storing all the action-reward pairs for every visited state, a task that incurs a huge memory overhead and complex computation. Instead, approximation methods have been proposed. Among the most successful ones is using a neural network in conjunction with RL, also known as deep RL. Deep models allow RL algorithms to solve complex problems in an end-to-end fashion, handle unstructured environments, learn complex functions, or predict actions in states that have not been visited in the past. Deep RL is gaining wide interest recently due to its success in robotics, Atari games, and superhuman capabilities~\cite{mnih2013playing,doya2000reinforcement,kober2013reinforcement,peters2003reinforcement}. Deep RL was the key technique behind defeating the human European champion in the game of Go, which has long been viewed as the most challenging of classic games for artificial intelligence~\cite{silver2016mastering}.

Many system optimization problems have a nature of delayed, sparse, aggregated or sequential rewards, where improving the long term sum of rewards is more important than a single immediate reward. For example, an RL environment can be a computer cluster. The state could be defined as a combination of the current resource utilization, available resources, time of the day, duration of jobs waiting to run, \textit{etc}. The action could be to determine on which resources to schedule each job. The reward could be the total revenue, jobs served in a time window, wait time, energy efficiency, \textit{etc}., depending on the objective. In this example, if the objective is to minimize the waiting time of all jobs, then a good solution must interact with the computer cluster and monitor the overall wait time of the jobs to determine good schedules. This behavior is inherent in RL. The RL agent has the advantage of not requiring expert labels or knowledge and instead the ability to learn directly from its own interaction with the world. RL can also learn sophisticated system characteristics that a straightforward solution like first come first served allocation scheme cannot. For instance, it could be better to put earlier long-running arrivals on hold if a shorter job requiring fewer resources is expected shortly. 

% Clearly specify the scope of the survey
In this paper, we review different attempts to overcome system optimization challenges with the use of deep RL. Unlike previous reviews~\cite{hameed2016survey,mahdavinejad2018machine,krauter2002taxonomy, wang2018machine,ashouri2018survey,luong2019applications} that focus on machine learning methods without discussing deep RL models or applying them beyond a specific system problem, we focus on deep RL in system optimization in general. From reviewing prior work, it is evident that standardized metrics for assessing deep RL solutions in system optimization problems are lacking. We thus propose quintessential metrics to guide future work in evaluating the use of deep RL in system optimization. We also discuss and address multiple challenges that faced when integrating deep RL into systems.  

 \vspace{-0.25cm}
\section{Background}

\label{bg}
% \begin{itemize}
%     \item description of markovian processes + every RL algorithm + equations + what is episodic and what is not
%     \item difference between model based and not
%     \item difference between off policy and on policy
%     \item difference between sarsa and Q-learning and explain what is temporal difference methods
%     \item multiarm bandit
% \end{itemize}
\begin{figure}
    \centering
    \includegraphics[width=0.5\textwidth]{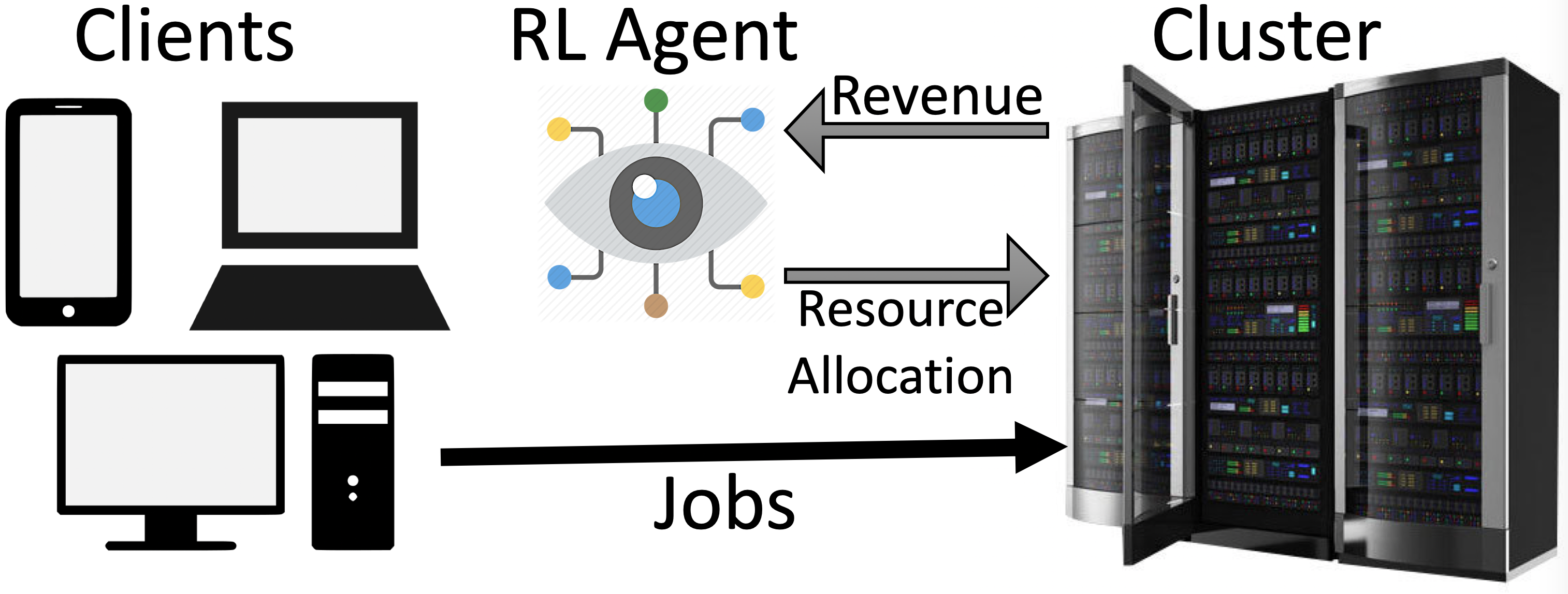}
    \caption{RL environment example. By observing the state of the environment (the cluster resources and arriving jobs' demands), the RL agent makes resource allocation actions for which he receives rewards as revenues. The agent's goal is to make allocations that maximize cumulative revenue.\vspace{-0.25cm}
}
    \label{fig:rl}
\end{figure}
One of the promising machine learning approaches is reinforcement learning (RL), in which an agent learns by continually interacting with an environment~\cite{kaelbling1996reinforcement}. In RL, the agent observes the state of the environment, and based on this state/observation takes an action as illustrated in figure~\ref{fig:rl}. The ultimate goal is to compute a policy--a mapping between the environment states and actions--that maximizes expected reward. 
RL can be viewed as a stochastic optimization solution for solving Markov Decision Processes (MDPs)~\cite{bellman1957}, when the MDP is not known. An MDP is defined by a tuple with four elements:
${S,A,P(s,a),r(s,a)}$ where $S$ is the set of states of the environment, $A$ describes the set of actions or transitions between states, $s' {\raise.17ex\hbox{$\scriptstyle\mathtt{\sim}$}}
 P(s,a)$ describes the probability distribution of next states given the current state and action and $r(s,a):S \times A \rightarrow R$ is the reward of taking action $a$ in state $s$. Given an MDP, the goal of the agent is to gain the largest possible cumulative reward. The objective of an RL algorithm associated with an MDP is to find a decision policy $\pi^*(a|s):s\rightarrow A$ that achieves this goal for that MDP: 
\begin{align}
\label{eq:MDP}
\begin{split}
    \pi^* = \argmaxA_\pi \mathbb{E}_{\tau{\raise.05ex\hbox{$\scriptstyle\mathtt{\sim}$}}\pi(\tau)} \left[\tau \right] =
    \\\argmaxA_\pi \mathbb{E}_{\tau{\raise.05ex\hbox{$\scriptstyle\mathtt{\sim}$}}\pi(\tau)} \left[\sum_{t}^{}r(s_t,a_t) \right],
\end{split}
\end{align}
where $\tau$ is a sequence of states and actions that define a single episode, and $T$ is the length of that episode.
Deep RL leverages a neural network to learn the policy (and sometimes the reward function). Over the past couple of years, a plethora of new deep RL techniques have been proposed~\cite{mnih2016asynchronous,ross2011,sutton2000,schulman2017proximal,lillicrap2015continuous}. 

Policy Gradient (PG)~\cite{sutton2000}, for example, uses a neural network to represent the policy. This policy is updated directly by differentiating the term in Equation~\ref{eq:MDP} as follows:
\begin{multline}
\label{eq:policygradient}
    \nabla_\theta J = \nabla_\theta\mathbb{E}_{\tau{\raise.05ex\hbox{$\scriptstyle\mathtt{\sim}$}}\pi(\tau)} \left[\sum_{t}^{}r(s_t,a_t) \right] 
    \\
    = \mathbb{E}_{\tau{\raise.05ex\hbox{$\scriptstyle\mathtt{\sim}$}}\pi(\tau)}\left[ \left(\sum_{t}^{}\nabla_\theta log\pi_\theta(a_{t}|s_{t})\right)\left(\sum_{t}^{}r(s_{t},a_{t})\right) \right]
    \\
    \approx \frac{1}{N}\sum_{i=1}^{N} \left[\left(\sum_{t}^{}\nabla_\theta log\pi_\theta(a_{i,t}|s_{i,t})\right)\left(\sum_{t}^{}r(s_{i,t},a_{i,t})\right) \right]
\end{multline}
and updating the network parameters (weights) in the direction of the gradient:
\begin{equation}
    \theta \leftarrow \theta+\alpha\nabla_\theta J,
\end{equation}
Proximal Policy Optimization (PPO)~\cite{schulman2017proximal} improves on top of PG for more deterministic, stable, and robust behavior by limiting the updates and ensuring the deviation from the previous policy is not large. 

In contrast, Q-Learning~\cite{watkins1992q}, state-action-reward-state-action (SARSA) \cite{rummery1994line} and deep deterministic policy gradient (DDPG)~\cite{lillicrap2015continuous} are temporal difference methods, \textit{i.e.}, they update the policy on every timestep (action) rather than on every episode. Furthermore, these algorithms bootstrap and, instead of using a neural network for the policy itself, they learn a Q-function, which estimates the long term reward from taking an action. The policy is then defined using this Q-function. 
In Q-Learning the Q-function is updated as follows:
\begin{multline}
    Q(s_t,a_t) \leftarrow Q(s_t,a_t) + r(s_t,a_t) + \gamma max_{a'_t}[Q(s'_t,a'_t)].
\end{multline}
In other words, the Q-function updates are performed based on the action that maximizes the value of that Q-function. On the other hand, in SARSA, the Q-function is updated as follows:
\begin{multline}
    Q(s_t,a_t) \leftarrow Q(s_t,a_t) + r(s_t,a_t) + \gamma Q(s_{t+1},a_{t+1}).
\end{multline}
In this case, the Q-function updates are performed based on the action that the policy would select given state $s_t$. DDPG fits multiple neural networks to the policy, including the Q-function and target time-delayed copies that slowly track the learned networks and greatly improve stability in learning.

Algorithms such as upper-confidence-bound and greedy can then be used to determine the policy based on the Q-function~\cite{auer2002using,sutton2018reinforcement}. The reviewed works in this paper focus on the epsilon greedy method where the policy is defined as follows:
\begin{equation}
        \pi^*(a_t|s_t)=
    \begin{cases}
      \argmaxA_{a_t} Q(s_t,a_t), & \text{w.p. }1-\epsilon
      \\
      random\ action, & \text{w.p. } \epsilon
    \end{cases}
\end{equation}

A method is considered to be on-policy if the new policy is computed directly from the decisions made by the current policy. PG, PPO, and SARSA are thus on-policy while DDPG and Q-Learning are off-policy. All the mentioned methods are model-free: they do not require a model of the environment to learn, but instead learn directly from the environment by trial and error. In some cases, a model of the environment could be available. It may also be possible to learn a model of the environment. This model could be used for planning and enable more robust training as less interaction with the environment may be required. 

Most RL methods considered in this review are structured around value function estimation (\textit{e.g.}, Q-values) and using gradients to update the policy. However, this is not always the case. For example, genetic algorithms, simulated annealing, genetic programming, and other gradient-free optimization methods - often called \textit{evolutionary} methods~\cite{sutton2018reinforcement} - can also solve RL problems in a manner analogous to the way biological evolution produces organisms with skilled behavior. Evolutionary methods can be effective if the space of policies is sufficiently small, the policies are common and easy to find, and the state of the environment is not fully observable. This review considers only the deep versions of these methods, \textit{i.e.}, using a neural network in conjunction with evolutionary methods typically used to evolve and update the neural network parameters or vice versa.

Multi-armed bandits~\cite{berry1985bandit,auer2002finite} simplify RL by removing the learning dependency on state and thus providing evaluative feedback that depends entirely on the action taken (1-step RL problems). The actions usually are decided upon in a greedy manner by updating the benefit estimates of performing each action independently from other actions. To consider the state in a bandit solution, contextual bandits may be used~\cite{chu2011contextual}. In many cases, a bandit solution may perform as well as a more complicated RL solution or even better. Many Bandit algorithms enjoy stronger theoretical guarantees on their performance even under adversarial settings.  These bounds would likely be of great value to the systems world as they suggest in the limit that the proposed algorithm would be no worse than using the best fixed system configuration in hindsight.  
\vspace{-0.25cm}
\subsection{Prior RL Works With Alternative Approximation Methods}
\vspace{-0.25cm}

Multiple prior works have proposed to use non-deep neural network approximation methods for RL in system optimization. These works include reliability and monitoring~\cite{das2014reinforcement,zhu2007reinforcement,zeppenfeld2008learning}, memory management~\cite{ipek2008self,andreasson2002collect,peled2015semantic,diegues2014self} in multicore systems, congestion control~\cite{li2016learning,silva2016smart}, packet routing~\cite{choi1996predictive,littman2013distributed,boyan1994packet}, algorithm selection~\cite{lagoudakis2000algorithm}, cloud caching~\cite{sadeghi2017optimal}, energy efficiency~\cite{farahnakian2014energy} and performance~\cite{peng2015random,jamshidi2015self,barrett2013applying,arabnejad2017comparison,mostafavi2018reinforcement}. Instead of using a neural network to approximate the policy, these works used tables, linear approximations, and other approximation methods to train and represent the policy. Tables were generally used to store the Q-values, \textit{i.e.}, one value for each action, state pair, which are used in training, and this table becomes the ultimate policy. In general, deep neural networks allowed for more complex forms of policies and Q functions~\cite{lin1993reinforcement}, and can better approximate good actions in new states.

 \vspace{-0.25cm}
\section{RL in System Optimization}

\label{problems}
% \usepackage{multirow}

In this section, we discuss the different system challenges tackled using RL and divide them into two categories: \textit{Episodic Tasks}, in which the agent-environment interaction naturally breaks down into a sequence of separate terminating episodes, and \textit{Continuing Tasks}, in which it does not. For example, when optimizing resources in the cloud, the jobs arrive continuously and there is not a clear termination state. But when optimizing the order of SQL joins, the query has a finite number of joins, and thus after enough steps the agent arrives at a terminating state.
\vspace{-0.25cm}
\subsection{Continuing Tasks}
\vspace{-0.25cm}

An important feature of RL is that it can learn from sparse reward signals, does not need expert labels, and the ability to learn direction from its own interaction with the world.
Jobs in the cloud arrive in an unpredictable and continuous manner. This might explain why many system optimization challenges tackled with RL are in the cloud~\cite{mao2016resource,he2017software,he2017integrated,tesauro2006online,xu2017deep,liu2017hierarchical,xu2012url,rao2009vconf}. A good job scheduler in the cloud should make decisions that are good in the long term. Such a scheduler should sometimes forgo short term gains in an effort to realise greater long term benefits. For example, it might be better to delay a long running job if a short running job is expected to arrive soon. The scheduler should also adapt to variations in the underlying resource performance and scale in the presence of new or unseen workloads combined with large numbers of resources.

These schedulers have a variety of objectives, including minimizing average performance of jobs and optimizing the resource allocation of virtual machines~\cite{mao2016resource,tesauro2006online,xu2012url,rao2009vconf}, optimizing data caching on edge devices and base stations ~\cite{he2017software,he2017integrated}, and maximizing energy efficiency~\cite{xu2017deep,liu2017hierarchical}. The RL algorithms used for addressing each system problem are listed in Table~\ref{tab:tab2} lists the RL algorithms used for addressing each problem. %The challenge in packet classification is similar to scheduling jobs in the cloud, in that the packets arrive continuously and in an unpredictable manner. \cite{liang2019neural} has thus leveraged deep RL to overcome this challenge.% Similarly, in multi-core systems where the state of the system changes over time, previous works suggested challenges at the memory level, namely, access patterns~\cite{ipek2008self}, prefetching~\cite{peled2015semantic}, garbage collection~\cite{andreasson2002collect} and transactional memory~\cite{diegues2014self}. Furthermore, for error recovery~\cite{zhu2007reinforcement}, and reliability~\cite{das2014reinforcement}. 

Interestingly, for cloud challenges most works are driven by Q-learning (or the very similar SARSA). In the absence of a complete environmental model, model-free Q-Learning can be used to generate optimal policies. It is able to make predictions incrementally by bootstrapping the current estimate with previous estimates and provide good sample efficiency~\cite{jin2018q}. Q-Learning is also characterized by inherent continuous temporal difference behavior where the policy can be updated immediately after each step (not the end of trajectory); something that might be very useful for online adaptation. %On the other hand PG based methods require episodic behavior, which is not the case for continuing tasks. Therefore, using PG methods in packet classification required special handling discussed in Section~\ref{subsec:dic}.%The bandit approach was not used mainly because it does not have observations and thus cannot generalize or adapt to new tasks.  

%For the multicore systems Q-learning might also be beneficial for similar reasons. However we observe a trend towards SARSA and Bandit. One reason behind this is the higher overhead of Q-learning which makes it prohibitive for some of these applications. For example, unlike the case in the cloud where a software implementation of the RL algorithm could be easily deployed, implementing a hardware prefetcher might affect the critical path and require more complicated hardware. Thus a bandit solution which is much simpler to implement might be more attractive. Similarly, SARSA requires less computation than Q-learning as it is on-policy and requires a single step in the policy to know the next step, unlike Q-learning that in every step requires to scan all the actions to find the one that maximizes the long term reward.
\vspace{-0.25cm}

\subsection{Episodic Tasks}
\vspace{-0.25cm}

Due to the sequential nature of decision making in RL, the order of the actions taken has a major impact on the rewards the RL agent collects. The agent can thus learn these patterns and select more rewarding actions. Previous works took advantage of this behavior in RL to optimize congestion control~\cite{jay2019deep,ruffy2018iroko}, decision trees for packet classification~\cite{liang2019neural}, sequence to SQL/program translation~\cite{zhong2017seq2sql,guu2017language,liang2016neural}, ordering of SQL joins~\cite{krishnan2018learning,ortiz2018learning,marcus2018deep,marcus2019neo}, compiler phase ordering~\cite{haj2019autophase,kulkarni2012mitigating} and device placement~\cite{addanki2019placeto,paliwal2019regal}.%, and algorithm selection in recursive functions~\cite{lagoudakis2000algorithm} (where different problem-size-dependant implementations of the same function could be called in a recursion).

After enough steps in these problems, the agent will always arrive at a clear
terminating step. For example, in query join order optimization, the number of joins is finite and known from the query. In congestion control -- where the routers need to adapt the sending rates to provide high throughput without comprising fairness -- the updates are performed on a fixed number of senders/receivers known in advance. These updates combined define one episode. This may explain why there is a trend towards using PG methods for these types of problems, as they don't require a continuous temporal difference behavior and can often operate in batches of multiple queries. Nevertheless, in some cases, Q-learning is still used, mainly for sample efficiency as the environment step might take a relatively long time. 

To improve the performance of PG methods, it is possible to take advantage of the way the gradient computation is performed. If the environment is not needed to generate the observation, it is possible to save many environment steps. This is achieved by rolling out the whole episode from interacting only with the policy and performing one environment step at the very end. The sum of rewards will be the same as the reward received from this environment step. For example, in query optimization, since the observations are encoded directly from the actions, and the environment is mainly used to generate the rewards, it will be possible to repeatedly perform an action, form the observation directly from this action, and feed it to the policy network. After the end of the episode, the environment can be triggered to get the final reward, which would be the sum of the intermediate rewards. This can significantly reduce the training time.

\vspace{-0.25cm}

\subsection{Discussion: Continuous vs. Episodic}
\vspace{-0.25cm}

\label{subsec:dic}
Continuous policies can handle both continuous and episodic tasks, while episodic policies cannot. So, for example, Q-Learning can handle all the tasks mentioned in this work, while PG based methods cannot directly handle it without modification. %For example, the authors in~\cite{liang2019neural} that used RL to derive building improved decision trees to classify packets, suggested to "take the somewhat unusual step of only computing rewards for the rollout when the tree is completed, and setting [the discount factor] $\gamma = 0$, effectively creating a series of 1-step decision problems similar to contextual bandits". Furthermore, these trees could extend to hundreds of thousands of nodes, requiring very long episodes (up to 15000 actions per episode) that complicate and prolong the training. In another 
For example, in~\cite{mao2016resource}, the authors limited the the scheduler window of jobs to $M$, allowing the agent in every time step to schedule up to $M$ jobs out of all arrived jobs. The authors also discussed this issue of "bounded time horizon" and hoped to overcome it by using a value network to replace the time-dependent baseline. It is interesting to note that prior work on continuous system optimization tasks using non deep RL approaches~\cite{choi1996predictive,littman2013distributed,boyan1994packet,peng2015random,jamshidi2015self,barrett2013applying,arabnejad2017comparison,sadeghi2017optimal,farahnakian2014energy} used Q-Learning.

One solution for handling continuing problems without episode boundaries with PG based methods is to define performance in terms of the average rate of reward per time step~\cite{sutton2018reinforcement} (Chapter 13.6). Such approaches can help better fit the continuous problems to episodic RL algorithms.
\begin{table*}[!t]
\caption{Problem formulation in the deep RL setting. The model abbreviations are: fully connected neural networks (FCNN), convolutional neural network (CNN), recurrent neural network (RNN), graph neural network (GNN), gated recurrent unit (GRU), and long short-term memory (LSTM).}
\label{tab:tab2}
\resizebox{1\textwidth}{!}{\begin{tabular}{|c|c|c|c|c|c|c|c|}
\hline
\textbf{Description} & \textbf{Reference} & \textbf{State/Observation} & \textbf{Action} & \textbf{Reward} & \textbf{Objective} &\textbf{Algorithm}& \textbf{Model} \\ \hline
%\textcolor{red}{\cite{boyan1994packet,choi1996predictive,littman2013distributed}}& Packet Routing &  & next node & \begin{tabular}[c]{@{}c@{}}transmission delay\\  + queueing time\end{tabular} & \begin{tabular}[c]{@{}c@{}}minimize network\\ congestion\end{tabular} & \begin{tabular}[c]{@{}c@{}}\textcolor{red}{Table}\end{tabular} \\ \hline
congestion control &\begin{tabular}[c]{@{}c@{}}\cite{jay2019deep}$^{1}$\\\cite{ruffy2018iroko}$^{2}$\end{tabular} &  \begin{tabular}[c]{@{}c@{}}histories of sending\\  rates and resulting\\  statistics (e.g., loss rate)\end{tabular} & changes to sending rate & \begin{tabular}[c]{@{}c@{}}throughput\\  and negative of\\  latency or\\ loss rate\end{tabular} & \begin{tabular}[c]{@{}c@{}}maximize throughput\\  while maintaining \\ fairness\end{tabular}& PPO$^{1,2}$/PG$^{2}$/DDPG$^{2}$ & FCNN \\ \hline
 packet classification &\cite{liang2019neural} & \begin{tabular}[c]{@{}c@{}}encoding of the\\ tree node, \textit{e.g.}, \\  split rules\end{tabular} & \begin{tabular}[c]{@{}c@{}}cutting a classification\\  tree node or partitioning\\  a set of rules\end{tabular} & \begin{tabular}[c]{@{}c@{}}classification time\\ /memory\\ footprint\end{tabular} & \begin{tabular}[c]{@{}c@{}}build optimal decision\\  tree for packet\\  classification\end{tabular}&PPO & FCNN \\ \hline
%\cite{lagoudakis2000algorithm} & Algorithm Selection & \begin{tabular}[c]{@{}c@{}}logscale of problem\\  size (e.g., number\\  of elements to sort)\end{tabular} & \begin{tabular}[c]{@{}c@{}}next algorithm\\  to use\end{tabular} & \begin{tabular}[c]{@{}c@{}}performance\\  improvement\end{tabular} & \begin{tabular}[c]{@{}c@{}}Minimize execution\\  time\end{tabular} & \begin{tabular}[c]{@{}c@{}}\textcolor{red}{Linear} \\ \textcolor{red}{Approximation}\\ \textcolor{red}{/Table}\end{tabular} \\ \hline
\begin{tabular}[c]{@{}c@{}}SQL join \\ order optimization\end{tabular} &\begin{tabular}[c]{@{}c@{}}\cite{krishnan2018learning}$^1$\\\cite{ortiz2018learning}$^2$\\\cite{marcus2019neo}$^3$\\\cite{marcus2018deep}$^4$\end{tabular} &   \begin{tabular}[c]{@{}c@{}}encoding of \\ current join plan\end{tabular} & next relation to join & \begin{tabular}[c]{@{}c@{}}negative cost$^{1-3}$,\\$1/cost^{4}$ \end{tabular}& \begin{tabular}[c]{@{}c@{}}minimize execution\\  time\end{tabular} &Q-Learning$^{1-3}$/PPO$^{4}$ &\begin{tabular}[c]{@{}c@{}}tree conv.$^3$,\\FCNN$^{1-4}$ \end{tabular}\\ \hline
%query optimization &\cite{marcus2019neo} &  query/plan encodings & next relation to join & negative cost & performance &Q-Learning& \begin{tabular}[c]{@{}c@{}}tree conv.,\\ FCNN \end{tabular}\\ \hline
% \begin{tabular}[c]{@{}c@{}}SQL join\\  order optimization\end{tabular} &\cite{marcus2018deep} & \begin{tabular}[c]{@{}c@{}}matrix encoding\\  of the join\\  tree structure\end{tabular} & next relation to join & 1/cost & \begin{tabular}[c]{@{}c@{}}minimize execution\\  time\end{tabular}&PPO & FCNN \\ \hline
\begin{tabular}[c]{@{}c@{}}sequence to\\  SQL\end{tabular}&\cite{zhong2017seq2sql} &   \begin{tabular}[c]{@{}c@{}}SQL vocabulary,\\  question, column\\  names\end{tabular} & \begin{tabular}[c]{@{}c@{}}query corresponding\\ to the token \end{tabular}& \begin{tabular}[c]{@{}c@{}}-2 invalid query, \\ -1 valid but wrong, \\ +1 valid and right\end{tabular} &
\begin{tabular}[c]{@{}c@{}}tokens in the\\ WHERE clause\end{tabular}&PG& LSTM \\ \hline
\begin{tabular}[c]{@{}c@{}}language to \\program translation \end{tabular} &\cite{guu2017language} &  \begin{tabular}[c]{@{}c@{}}natural language\\ utterances\end{tabular} & \begin{tabular}[c]{@{}c@{}} a sequence of \\program tokens\end{tabular} & \begin{tabular}[c]{@{}c@{}}1 if correct result\\
0 otherwise\end{tabular}& \begin{tabular}[c]{@{}c@{}}generate equivalent \\ program \end{tabular}&PG& \begin{tabular}[c]{@{}c@{}}LSTM,\\ FCNN \end{tabular}\\ \hline
semantic parsing &\cite{liang2016neural} &  \begin{tabular}[c]{@{}c@{}} embedding of \\ the words \end{tabular} & \begin{tabular}[c]{@{}c@{}} a sequence of \\program tokens\end{tabular} & \begin{tabular}[c]{@{}c@{}}positive if correct\\
0 otherwise\end{tabular} & \begin{tabular}[c]{@{}c@{}}generate equivalent \\ program \end{tabular} &PG& \begin{tabular}[c]{@{}c@{}}RNN,\\ GRU \end{tabular} \\ \hline
%\cite{das2014reinforcement} & \begin{tabular}[c]{@{}c@{}}reliability \&\\ monitoring\end{tabular}  & \begin{tabular}[c]{@{}c@{}}thermal aging \& \\ stress\end{tabular}  &\begin{tabular}[c]{@{}c@{}} CPU frequency \& \\ thread affinity\end{tabular} &  &  & \textcolor{red}{table} \\ \hline
%\cite{peled2015semantic,ipek2008self,zhu2007reinforcement} &  &  &  &  &  & \textcolor{red}{table} \\ \hline
%\cite{andreasson2002collect} &  &  &  &  &  & \textcolor{red}{tile coding} \\ \hline
%\cite{diegues2014self} &  &  &  &  &  & \textcolor{red}{not deep} \\ \hline
\begin{tabular}[c]{@{}c@{}}resource allocation \\in the cloud \end{tabular} &\cite{mao2016resource} &  \begin{tabular}[c]{@{}c@{}}current allocation of \\cluster resources \&\\ resource profiles of\\ waiting jobs\end{tabular} & \begin{tabular}[c]{@{}c@{}}next job \\ to schedule\end{tabular} &\begin{tabular}[c]{@{}c@{}} $\Sigma_i(\frac{-1}{T_i})$ for \\all jobs in the\\ system ($T_i$ is the\\ duration of job $i$)\end{tabular} &\begin{tabular}[c]{@{}c@{}} minimize average \\job slowdown\end{tabular} &PG &FCNN \\ \hline
%\cite{sadeghi2017optimal} &  &  &  &  &  & \begin{tabular}[c]{@{}c@{}}\textcolor{red}{linear} \\ \textcolor{red}{approximation}\end{tabular}\\ \hline

resource allocation  &\cite{he2017software,he2017integrated} &  \begin{tabular}[c]{@{}c@{}}status of edge\\ devices, base stations,\\ content caches\end{tabular}  &\begin{tabular}[c]{@{}c@{}} which base station,\\ to offload/cache\\ or not\end{tabular}  & total revenue & \begin{tabular}[c]{@{}c@{}} maximize total\\ revenue\end{tabular} &Q-Learning& CNN\\ \hline
%\cite{peng2015random} &  &  &  &  & \begin{tabular}[c]{@{}c@{}}performance/\\waiting time\end{tabular} & \textcolor{red}{table} \\ \hline
\begin{tabular}[c]{@{}c@{}}resource allocation \\in the cloud \end{tabular} &\cite{tesauro2006online} &  \begin{tabular}[c]{@{}c@{}}current allocation \\ \& demand\end{tabular} & \begin{tabular}[c]{@{}c@{}}next resource\\ to allocate\end{tabular}  & payments & maximize revenue & Q-Learning&FCNN\\ \hline
%\cite{farahnakian2014energy} &  &  &  &  &  & \textcolor{red}{\begin{tabular}[c]{@{}c@{}}table\\/not deep\end{tabular}}\\ \hline
\begin{tabular}[c]{@{}c@{}}resource allocation \\in cloud radio \\access networks \end{tabular} &\cite{xu2017deep} &  \begin{tabular}[c]{@{}c@{}}active remote radio \\ heads \& user demands \end{tabular} & \begin{tabular}[c]{@{}c@{}}which remote \\radio heads\\ to activate  \end{tabular}& \begin{tabular}[c]{@{}c@{}}negative power\\ consumption\end{tabular} & \begin{tabular}[c]{@{}c@{}}power\\ efficiency\end{tabular} &Q-Learning& FCNN \\ \hline
\begin{tabular}[c]{@{}c@{}}cloud resource\\
allocation \& \\power management\end{tabular} &\cite{liu2017hierarchical} &   \begin{tabular}[c]{@{}c@{}}current allocation \\ \& demand\end{tabular} & \begin{tabular}[c]{@{}c@{}}next resource\\ to allocate\end{tabular} & \begin{tabular}[c]{@{}c@{}}linear combination \\ of total power
,\\ VM latency, \& \\  reliability metrics \end{tabular} & power efficiency &Q-Learning& \begin{tabular}[c]{@{}c@{}}autoencoder,\\ weight sharing\\ \& LSTM\end{tabular} \\ \hline
%\cite{jamshidi2015self,arabnejad2017comparison} &  &  &  &  &  & \begin{tabular}[c]{@{}c@{}}\textcolor{red}{table}\\/fuzzy Q-learn \end{tabular}\\ \hline
%\cite{barrett2013applying} &  &  &  &  &  & \textcolor{red}{table}\\ \hline
%\cite{rao2009vconf,xu2012url} & \begin{tabular}[c]{@{}c@{}}automate virtual \\machine (VM)\\ configuration process\end{tabular} &  &  &  & \begin{tabular}[c]{@{}c@{}} optimal system\\
%wide performance \end{tabular}& FCNN\\ \hline
\begin{tabular}[c]{@{}c@{}}automate virtual \\machine (VM)\\ configuration process\end{tabular} &\begin{tabular}[c]{@{}c@{}}\cite{rao2009vconf}\\\cite{xu2012url}\end{tabular} &  \begin{tabular}[c]{@{}c@{}}current resource\\ allocations\end{tabular} &\begin{tabular}[c]{@{}c@{}} increase/decrease \\ CPU/time/memory\end{tabular}&\begin{tabular}[c]{@{}c@{}}throughput\\-response time\end{tabular}  & maximize performance &Q-Learning& \begin{tabular}[c]{@{}c@{}}FCNN,\\ model-based\end{tabular}\\ \hline
\begin{tabular}[c]{@{}c@{}}compiler phase\\  ordering\end{tabular} &\begin{tabular}[c]{@{}c@{}}\cite{kulkarni2012mitigating}$^1$\\\cite{haj2019autophase}$^2$\end{tabular} &  program features & \begin{tabular}[c]{@{}c@{}}next optimization\\  pass\end{tabular} & \begin{tabular}[c]{@{}c@{}}performance\\  improvement\end{tabular} & \begin{tabular}[c]{@{}c@{}}minimize execution\\  time\end{tabular} &\begin{tabular}[c]{@{}c@{}} Evolutionary Methods$^1$/\\ Q-Learning$^2$/PG$^2$\end{tabular}& FCNN \\ \hline
%\begin{tabular}[c]{@{}c@{}}Compiler Phase\\  Ordering\end{tabular}& \cite{kulkarni2012mitigating}&program features &netx optimization  & & & FCNN\\ \hline

device placement&\begin{tabular}[c]{@{}c@{}}\cite{paliwal2019regal}$^1$\\\cite{addanki2019placeto}$^2$\end{tabular} &computation graph & \begin{tabular}[c]{@{}c@{}}placement/schedule\\ of graph node\end{tabular} & speedup & \begin{tabular}[c]{@{}c@{}}maximize performance\\ \& minimize peak\\ memory\end{tabular}&\begin{tabular}[c]{@{}c@{}}PG$^{1,2}$/\\ Evolutionary Methods$^1$\end{tabular}& GNN/FCNN\\ \hline
%device placement&\cite{paliwal2019regal} &computation graph & \begin{tabular}[c]{@{}c@{}}placement/schedule\\ of graph node\end{tabular}& & & GNN/FCNN\\ \hline
\begin{tabular}[c]{@{}c@{}}distributed instr-\\uction  placement\end{tabular}&\cite{coons2008feature} & instruction features & \begin{tabular}[c]{@{}c@{}}instruction placement\\ location\end{tabular}& speedup& maximize performance & Evolutionary Methods&FCNN \\ \hline
\end{tabular}}
\vspace{-0.5cm}

\end{table*}

\begin{table*}[!t]
\caption{Evaluation results.  }
\label{tab:results}
\resizebox{1\textwidth}{!}{\begin{tabular}{|c|c|c|c|c|c|c|c|}
\hline
\textbf{Work} & \textbf{Problem} & \textbf{\begin{tabular}[c]{@{}c@{}}Environment \\ Step Time\end{tabular}} & \textbf{\begin{tabular}[c]{@{}c@{}}Number of Steps \\ Per Iteration\end{tabular}} & \textbf{\begin{tabular}[c]{@{}c@{}}Number of training \\ Iterations\end{tabular}} & \textbf{\begin{tabular}[c]{@{}c@{}}Total Number \\ Of Steps\end{tabular}} & \textbf{\begin{tabular}[c]{@{}c@{}}Improves State\\ of the Art\end{tabular}} & \textbf{\begin{tabular}[c]{@{}c@{}}Compares Against\\ Bandit/Random \\ Search\end{tabular}} \\ \hline
 \begin{tabular}[c]{@{}c@{}}packet \\ classification\end{tabular} &\cite{liang2019neural} & 20-600ms & up to 60,000 & up to 167 & 1,002,000 & \checkmark (18\%) & \ding{53} \\ \hline
\begin{tabular}[c]{@{}c@{}}congestion \\ control\end{tabular} &\cite{jay2019deep} &  50-500ms & 8192 & 1200 & 9,830,400 & \checkmark (similar) & \checkmark \\ \hline
\begin{tabular}[c]{@{}c@{}}congestion \\ control\end{tabular} &\cite{ruffy2018iroko} &  0.5s & N/A & N/A & 50,000-100,000 & \ding{53} & \ding{53} \\ \hline
\begin{tabular}[c]{@{}c@{}}resource\\ allocation\end{tabular} &\cite{mao2016resource} &  10-40ms & 20,000 & 1000 & 20,000,000 & \checkmark (10-63\%) & \checkmark \\ \hline
\begin{tabular}[c]{@{}c@{}}resource\\ allocation\end{tabular} &\begin{tabular}[c]{@{}c@{}}\cite{he2017software}\\\cite{he2017integrated}\end{tabular} &  N/A & N/A & 20,000 & N/A & no comparison & \ding{53} \\ \hline
\begin{tabular}[c]{@{}c@{}}resource\\ allocation\end{tabular} &\cite{tesauro2006online} &  N/A & N/A & 10,000-20,000 & N/A & no comparison & \checkmark \\ \hline
\begin{tabular}[c]{@{}c@{}}resource\\ allocation\end{tabular} &\cite{xu2017deep} &  N/A & N/A & N/A & N/A & no comparison & \checkmark \\ \hline
 \begin{tabular}[c]{@{}c@{}}resource\\ allocation\end{tabular} &\cite{liu2017hierarchical} & 1-120 minutes & 100,000 & 20 & 2,000,000 & no comparison & \ding{53} \\ \hline
\begin{tabular}[c]{@{}c@{}}resource\\ allocation\end{tabular} &\begin{tabular}[c]{@{}c@{}}\cite{rao2009vconf}\\\cite{xu2012url}\end{tabular} &  N/A & N/A & N/A & N/A & no comparison & \ding{53} \\ \hline
\begin{tabular}[c]{@{}c@{}}SQL\\ Joins\end{tabular} &\cite{krishnan2018learning} &  ~10ms & ~640 & ~100 & ~64,000 & \checkmark (70\%) & \checkmark \\ \hline
\begin{tabular}[c]{@{}c@{}}SQL\\ joins\end{tabular} &\cite{ortiz2018learning} &  N/A & N/A & N/A & N/A & no comparison & \ding{53} \\ \hline
\begin{tabular}[c]{@{}c@{}}SQL\\ joins\end{tabular} &\cite{marcus2019neo} & 250ms & 100-8,000 & 100 & 10,000-80,000 & \checkmark (10-66\%) & \checkmark \\ \hline
\begin{tabular}[c]{@{}c@{}}SQL\\ joins\end{tabular} &\cite{marcus2018deep} &  1.08s & N/A & N/A & 10,000 & \checkmark (20\%) & \checkmark \\ \hline
\begin{tabular}[c]{@{}c@{}} sequence to\\ SQL\end{tabular} &\cite{zhong2017seq2sql} & N/A & 80,654 & 300 & 24,196,200 & \checkmark (similar) & \ding{53} \\ \hline
\begin{tabular}[c]{@{}c@{}}language to \\ program trans.\end{tabular} &\cite{guu2017language} &  N/A & N/A & N/A & 13,000 & \checkmark (56\%) & \ding{53} \\ \hline
\begin{tabular}[c]{@{}c@{}}semantic\\ parsing\end{tabular} &\cite{liang2016neural} &  N/A & 3,098 & 200 & 619,600 & \checkmark (3.4\%) & \ding{53} \\ \hline
\begin{tabular}[c]{@{}c@{}}phase \\ ordering\end{tabular} &\cite{haj2019autophase} &  ~1s & N/A & N/A & 1,000-10,000 & \checkmark (similar) & \checkmark \\ \hline
\begin{tabular}[c]{@{}c@{}}phase \\ ordering\end{tabular} &\cite{kulkarni2012mitigating}& \begin{tabular}[c]{@{}c@{}} 13.2 days\\ for all steps\end{tabular} &N/A &N/A &N/A & \ding{53} & \ding{53}\\ \hline
\begin{tabular}[c]{@{}c@{}}device \\ placement \end{tabular} &\cite{addanki2019placeto} & N/A (seconds) &N/A &N/A &1,600-94,000 &\checkmark (3\%) &\checkmark \\ \hline
\begin{tabular}[c]{@{}c@{}}device\\ placement \end{tabular} &\cite{paliwal2019regal} & N/A (seconds)&N/A &N/A &400,000 & \checkmark (5\%) &\checkmark \\ \hline
\begin{tabular}[c]{@{}c@{}} instruction \\ placement\end{tabular} &\cite{coons2008feature} &N/A (minutes) &N/A & 200 &N/A (days) &\ding{53} &\ding{53} \\ \hline

\end{tabular}}
\vspace{-0.5cm}

\end{table*}

\vspace{-0.25cm}
\section{Formulating the RL environment}

\label{model}
Table~\ref{tab:tab2} lists all the works we reviewed and their problem formulations in the context of RL, \textit{i.e.}, the model, observations, actions and rewards definitions. Among the major challenges when formulating the problem in the RL environment is properly defining the system problem as an RL problem, with all of the required inputs and outputs, \textit{i.e.}, state, action spaces and rewards. The rewards are generally sparse and behave similarly for different actions, making the RL training ineffective due to bad gradients. The states are generally defined using hand engineered features that are believed to encode the state of the system. This results in a large state space with some features that are less helpful than others and rarely captures the actual system state. Using model-based RL can alleviate this bottleneck and provide more sample efficiency. \cite{liu2017hierarchical} used auto-encoders to help reduce the state dimensionality. The action space is also large but generally represents actions that are directly related to the objective. Another challenge is the environment step. Some tasks require a long time for the environment to perform one step, significantly slowing the learning process of the RL agent.

Interestingly, most works focus on using simple out-of-the-box FCNNs, while some works that targeted parsing and translation (\cite{liang2016neural,guu2017language,zhong2017seq2sql}) used RNNs~\cite{graves2013speech} due to their ability to parse strings and natural language. While FCNNs are simple and easy to train to learn a linear and non-linear function policy mappings, sometimes having a more complicated network structure suited for the problem could further improve the results.

\vspace{-0.25cm}
\subsection{Evaluation Results}
\vspace{-0.25cm}

Table~\ref{tab:results} lists training, and evaluation results of the reviewed works. We consider the time it takes to perform a step in the environment, the number of steps needed in each iteration of training, number of training iterations, total number of steps needed, and whether the prior work improves the state of the art and compares against random search/bandit solution. 

The total number of steps and the the cost of each environment step is important to understand the sample efficiency and practicality of the solution, especially when considering RL’s inherent sample inefficiency~\cite{schaal1997learning,hester2018deep}. For different workloads, the number of samples needed varies from thousands to millions. The environment step time also varies from milliseconds to minutes. In multiple cases, the interaction with the environment is very slow. Note that in most cases when the environment step time was a few milliseconds, it was because it was a simulated environment, not a real one. We observe that for faster environment steps more training samples were gathered to leverage that and further improve the performance. This excludes \cite{liu2017hierarchical} where a cluster was used and thus more samples could be processed in parallel.

As listed in Table~\ref{tab:results}, many works did not provide sufficient data to reproduce the results. Reproducing the results is necessary to further improve the solution and enable future evaluation and comparison against it.
\vspace{-0.25cm}
\subsection{Frameworks and Toolkits}
\vspace{-0.25cm}

A few RL benchmark toolkits for developing and comparing reinforcement learning algorithms, and providing a faster simulated system environment, were recently proposed. OpenAI Gym~\cite{brockman2016openai} supports an environment for teaching agents everything, from walking to playing games like Pong or Pinball. Iroko~\cite{ruffy2018iroko} provides a data center emulator to understand the requirements and limitations of applying RL in data center networks. It interfaces with the OpenAI Gym and offers a way to evaluate centralized and decentralized RL algorithms against conventional traffic control solutions.

Park~\cite{mao2019park} proposes an open platform for easier formulation of the RL environment for twelve real world system optimization problems with one common easy to use API. The platform provides a translation layer between the system and the RL environment making it easier for RL researchers to work on systems problems. That being said, the framework lacks the ability to change the action, state and reward definitions, making it harder to improve the performance by easily modifying these definitions. 

%\subsection{Discussion: Model Based RL}
 \section{Considerations for Evaluating Deep RL in System Optimization}

\label{metrics}
In this section, we propose a set of questions that can help system optimization researchers determine whether deep RL could be an effective tool in solving their systems optimization challenges.
\vspace{-0.25cm}

\subsection*{Can the System Optimization Problem Be Modeled by an MDP?}
\vspace{-0.25cm}

The problem of RL is the optimal control of an MDP. MDPs are a classical formalization of sequential decision making, where actions influence not just immediate rewards but also future states and rewards. This involves delayed rewards and the trade-off between delayed and immediate reward. In MDPs, the new state and new reward are dependent only on the preceding state and action. Given a perfect model of the environment, an MDP can compute the optimal policy. 

MDPs are typically a straightforward formulation of the system problem, as an agent learns by continually interacting with the system to achieve a particular goal, and the system responds to these interactions with a new state and reward. The agent's goal is to maximize expected reward over time.
\vspace{-0.25cm}

\subsection*{Is It a Reinforcement Learning Problem?}
\vspace{-0.25cm}

What distinguishes RL from other machine learning approaches is the presence of self exploration and exploitation, and the tradeoff between them.  For example, RL is different from supervised learning. The latter is learning from a training set with labels provided by an external supervisor that is knowledgeable. For each example the label is the correct action the system should take. The objective of this kind of learning is to act correctly in new situations not present in the training set. However, supervised learning is not suitable for learning from interaction, as it is often impractical to obtain examples representative of all the cases in which the agent has to act.

\vspace{-0.25cm}

\subsection*{Are the Rewards Delayed?}
\vspace{-0.25cm}

RL algorithms do not maximize the immediate reward of taking actions but, rather, expected reward over time. For example, an RL agent can choose to take actions that give low immediate rewards but that lead to higher rewards overall, instead of taking greedy actions every step that lead to high immediate rewards but low rewards overall. If the objective is to maximize the immediate reward or the actions are not dependent, then other simpler approaches, such as bandits and greedy algorithms, will perform better than deep RL, as their objective is to maximize the immediate reward.

\vspace{-0.25cm}

\subsection*{What is Being Learned?}
\vspace{-0.25cm}

It is important to provide insights on what is being learned by the agent. For example, what actions are taken in which states and why? Can the knowledge learned be applied to new states/tasks? Is there a structure to the problem being learned? If a brute-force solution is possible for simpler tasks, it will also be helpful to know how much better the performance of the RL agent is than the brute force solution. In some cases, not all hand-engineered features are useful. Using all of them can result in high variance and prolonged training. Feature analysis can help overcome this challenge. For example, in~\cite{coons2008feature} significant performance gaps were shown for different feature selection.  
\vspace{-0.25cm}

\subsection*{Does It Outperform Random Search and a Bandit Solution?}
\vspace{-0.25cm}

 In some cases, the RL solution is just another form of a improved random search. In some cases, good RL results were achieved merely by chance. For instance, if the features used to represent the state are not good or do not have a pattern that could be learned. In such cases, random search might perform as well as RL, or even better, as it is less complicated. For example, in \cite{haj2019autophase}, the authors showed 10\% improvement over the baseline by using random search. In some cases the actions are independent and a greedy or bandit solution can achieve the optimal or near-optimal solution. Using a bandit method is equivalent using a 1-step RL solution, in which the objective is to maximize the immediate reward. Maximizing the immediate reward could deliver the overall maximum reward and, thus, a comparison against a bandit solution can help reveal this.
 \vspace{-0.25cm}

 \subsection*{Are the Expert Actions Observable?}
 \vspace{-0.25cm}

In some cases it might be possible to have access to \textit{expert actions}, \textit{i.e.}, optimal actions. For example, if a brute force search is plausible and practical then it is possible to outperform deep RL by using it or using imitation learning~\cite{schaal1999imitation}, which is a supervised learning approach that learns by imitating expert actions.
\vspace{-0.25cm}

 \subsection*{Is It Possible to Reproduce/Generalize Good Results?}
 \vspace{-0.25cm}

 The learning process in deep RL is stochastic and thus good results are sometimes achieved due to local maxima, simple tasks, and chance. In~\cite{haarnoja2018soft} different results were generated by just changing the random seeds. In many cases, good results cannot be reproduced by retraining, training on new tasks, or generalizing to new tasks. 

\vspace{-0.25cm}

\subsection*{Does It Outperform the State of the Art?}
\vspace{-0.25cm}

The most important metric in the context of system optimization in general is outperforming the state of the art. Improving the state of the art includes different objectives, such as efficiency, performance, throughput, bandwidth, fault tolerance, security, utilization, reliability, robustness, complexity, and energy. If the proposed approach does not perform better than the state of the art in some metric then it is hard to justify using it. Frequently, the state of the art solution is also more stable, practical, and reliable than deep RL. In many prior works listed in Table~\ref{tab:results} a comparison against the state of the art is not available or deep RL performs worse. In some cases deep RL can perform as good as the state of the art or slightly worse, but still be a useful solution as it achieves an improvement on other metrics.

 \section{RL Methods and Neural Network Models}

Multiple RL methods and neural network models can be used. RL frameworks like RLlib~\cite{liang2017ray}, Intel's Coach~\cite{caspi_itai_2017_1134899}, TensorForce~\cite{tensorforce}, Facebook Horizon~\cite{gauci2018horizon}, and Google's Dopamine~\cite{castro18dopamine} can help the users pick the right RL model, as they provide implementations of many policies and models for which a convenient interface is available.

As a rule of thumb, we rank RL algorithms based on sample efficiency as follows: model-based approaches (most efficient), temporal difference methods, PG methods, and evolutionary algorithms (least efficient). In general, many RL environments run in a simulator. For example~\cite{paliwal2019regal,mao2019park,mao2016resource}, run in a simulator as the real environment's step would take minutes or hours, which significantly slows down the training. If this simulator is fast enough or training time is not constrained then PG methods can perform well. If the simulator is not fast enough or training time is constrained then temporal difference methods can do better than PG methods as they are more sample efficient. 

If the environment is the real one, then temporal difference can do well, as long as interaction with the environment is not slow.  Model-based RL performs better if the environment is slow. Model-based methods require a model of the environment (that can be learned) and rely mainly on planning rather than learning~\cite{deisenroth2011learning,guo2014deep}. Since planning is not done in the actual environment, but in much faster simulation steps within the model, it requires less samples from the real environment to learn. Many real-world system problems have well established and often highly accurate models, which model-based methods can leverage. 
That being said, model-free methods are often used as they are simpler to deploy and have the potential to generalize better from exploration in a real environment.

If good policies are easy to find and if either the space of policies is small enough or time is not a bottleneck for the search, then evolutionary methods can be effective. Evolutionary methods also have advantages when the learning agent cannot observe the complete state of the environment. As mentioned earlier, bandit solutions are good if the problem can be viewed as a one-step RL problem.

PG methods are in general more stable than methods like Q-Learning that do not directly use and derive a neural network to represent the agent's policy. The greedy nature of directly deriving the policy and moving the gradient in the direction of the objective also make PG methods easier to reason about and often more reliable. However, Q-Learning can be applied to data collected from a running system more readily than PG, which must interact with the system during training.

%\textcolor{red}{TODOOO: TED ADD HERE PLEASE when to use each neural network topology. For example, CNNS for image observations, etc.}

The RL methods may be implemented using any number of deep neural network architectures.  The preferred architecture depends on the the nature of the observation and action spaces.  CNNs that efficiently capture spatially-organized observation spaces lend themselves visual data (\textit{e.g.}, images or video).  Networks designed for sequential learning, such as RNNs, are appropriate for  observation spaces involving sequence data (\textit{e.g.}, code, queries, temporal event streams). Otherwise, FCNNs are preferred for their general applicability and ease of use, although they tend to be the most computationally-intensive choice.  Finally, GNNs or other networks that capture structure within observations can be used in the less frequent case that the designer has a priori knowledge of the representational structure.  In this case, the model can even generate structured action spaces (\textit{e.g.}, a query plan tree or computational graph).

 \vspace{-0.25cm}
\section{Challenges}

\label{challenges}
In this section, we discuss the primary challenges that face the application of deep RL in system optimization.
%\textbf{Using The Right RL Model} is necessary to achieve good quality of results. This includes using appropriate agents and RL algorithms that fit the problem. For example, %to use episodic policies with episodic tasks and continuous policies with continuous tasks. In addition, 
%to use proper embedding of the states that have meaningful features that directly reflect upon the rewards and actions. Furthermore, proper neural network structures and appropriate hyperparameter search are necessary to guarantee robust policy approximation. RL frameworks like RLlib~\cite{liang2017ray}, Intel's Coach, TensorForce, Facebook Horizon~\cite{gauci2018horizon}, and Google's Dopamine can help the users pick the right RL model as they allow convenient use of different RL algorithms and customized neural network models

% \paragraph{Interactions with Real Systems can be Slow.} 
% Rewrote the title, check this 

\textbf{Interactions with Real Systems Can Be Slow. Generalizing from Faster Simulated Environments Can Be Restrictive.}
Unlike the case with simulated environments that can run fast, when running on a real system, performing an action can trigger a reward after a lengthy delay. For example, when scheduling jobs on a cluster of nodes, some jobs might require hours to run, and thus improving their performance by monitoring job execution time will be very slow. To speed up this process, some works use simulators as cost models instead of the actual system. These simulators often do not fully capture the actual behavior of the real system and thus the RL agent may not work as well in practice. More comprehensive environment models can aid generalization from simulated environments. RL methods that are more sample efficient will speed up training in real system environments.

\textbf{Instability and High Variance.} This is a common problem which leads to bad policies when tackling system problems with deep RL. Such policies can generate a large performance gap when trained multiple times and behave in an unpredictable manner. This is mainly due to poor formulation of the problem as an RL problem, limited observation of the state, \textit{i.e.,} the use of embeddings and input features that are not sufficient/meaningful, and sparse or similar rewards. Sparse rewards can be due to bad reward definition or the fact that some rewards cannot be computed directly and are known only at the end of the episode. For example, in \cite{liang2019neural}, where deep RL is used to optimize decision trees for packet classification, the reward (the performance of the tree) is known only when the whole tree is built, or after approximately 15,000 steps. In some cases using more robust and stable policies can help. For example, Q-learning is known to have good sample efficiency but unstable behavior. SARSA, double Q-learning~\cite{van2016deep} and policy gradient methods, on the other hand, are more stable. Subtracting a bias in PG can also help reduce variance~\cite{greensmith2004variance}. 

\textbf{Lack of Reproducibility.} Reproducibility is a frequent challenge with many recent works in system optimization that rely on deep RL. It becomes difficult to reproduce the results due to restricted access to the resources, code, and workloads used, lack of a detailed list of the used network hyperparameters and lack of stable, predictable, and scalable behavior of the different RL algorithms. This challenge prevents future deployment, incremental improvements, and proper evaluation.

\textbf{Defining Appropriate Rewards, Actions and States.} The proper definition of states, actions, and rewards is the key, since otherwise the RL solution is not useful. In the general use case of deep RL, defining the states, actions and rewards is much more straightforward than in the case in system optimization. For example, in atari games, the state is an image representing the current status of the game, the rewards are the points collected while playing and the actions are moves in the game. However, often in system optimization, it is not clear what are the appropriate definitions. Furthermore, in many cases the rewards are sparse or similar, the states are not fully observable to capture the whole system state and have limited features that capture only a small portion of the system state. This results in unstable  and inadequate policies. Generally, the action and state spaces are large, requiring a lot of samples to learn and resulting in instability and large variance in the learned network. Therefore, retraining often fails to generate the same results.

\textbf{Lack of Generalization.} The lack of generalization is an issue that deep RL solutions often suffer from. This might be beneficial when learning a particular structure. For example, in NeuroCuts~\cite{liang2019neural}, the target is to build the best decision tree for fixed set of predefined rules and thus the objective of the RL agent is to find the optimal fit for these rules. However, lack of generalization sometimes results in a solution that works for a particular workload or setting but overall, across various workloads, is not very good. This problem manifests when generalization is important and the RL agent has to deal with new states that it did not visit in the past. For example, in~\cite{paliwal2019regal,addanki2019placeto}, where the RL agent has to learn good resource placements for different computation graphs, the authors avoided the possibility of learning only good placements for particular computation graphs by training and testing on a wide range graphs.

\textbf{Lack of Standardized Benchmarks, Frameworks and Evaluation Metrics.} The lack of standardized benchmarks, frameworks and evaluation metrics makes it very difficult to evaluate the effectiveness of the deep RL methods in the context of system optimization. Thus, it is crucial to have proper standardized frameworks and evaluation metrics that define success. Moreover, benchmarks are needed that enable proper training, evaluation of the results, measuring the generalization of the solution to new problems and performing valid comparisons against baseline approaches.

 \vspace{-0.25cm}
\section{An Illustrative Example}
\vspace{-0.25cm}

%Ameer, this looks good, it might make it even better if you add some paragraph titles to highlight the different components.
We put all the metrics (from Section~\ref{metrics}) to work and further highlight the challenges (from Section~\ref{challenges}) of implementing deep RL solutions using DeepRM~\cite{mao2016resource} as an illustrative example. In DeepRM, the targeted system problem is resource allocation in the cloud. The objective is to avoid job slowdown, \textit{i.e.}, the goal is to minimize the wait time for all jobs. DeepRM uses PG in conjunction with a simulated environment rather than a real cloud environment. This significantly improves the step time but can result in restricted generalization when used in a real environment. Furthermore, since all the simulation parameters are known, the full state of the simulated environment can be captured. The actions are defined as selecting which job should be scheduled next. The state is defined as the current allocation of cluster resources, as well as the resource profiles of jobs waiting to be scheduled. The reward is defined as the sum of of job slowdowns: $\Sigma_i(\frac{-1}{T_i})$ where $T_i$ is the pure execution time of job $i$ without considering the wait time. This reward basically gives a penalty of $-1$ for jobs that are waiting to be scheduled. The penalty is divided by $T_i$ to give a higher priority to shorter jobs.

\emph{The state, actions and reward clearly define an MDP} and \emph{a reinforcement learning problem}. Specifically, the agent interacts with the system by making sequential allocations, observing the state of the current allocation of resources and receiving delayed long-term rewards as overall slow downs of jobs.
\emph{The rewards are delayed} because the agent cannot know the effect of the current allocation action on the overall slow down at any particular time step; the agent would have to wait until all the other jobs are allocated to assess the full impact. The agent then \emph{learns which jobs to allocate in the current time step to minimize the average job slowdown, given the current resource allocation in the cloud}. Note that DeepRM also learns to withhold larger jobs to make room for smaller jobs to reduce the overall average job slowdown. DeepRM is shown to \emph{outperform random search}. %The authors also show that DeepRM learns to withhold large jobs to make room for small jobs and reduce the overall average job slowdown. In addition, DeepRM is shown to outperform random search.

\emph{Expert actions are not available in this problem} as there are no methods to find the optimal allocation decision at any particular time step. During training in DeepRM, \emph{multiple examples of job arrival sequences were considered to encourage policy generalization and robust decisions}\footnote{\label{note1}Results provided were only in the simulated system.}. DeepRM is also shown to \emph{outperform the state-of-the-art by $10$--$63$\%}$^1$.

Clearly, in the case of DeepRM, most of the challenges mentioned in Section~\ref{challenges} are manifested. \emph{The interaction with the real cloud environment is slow} and thus the authors opted for a simulated environment. This has the advantage of speeding up the training but may result in a policy that does not generalize to the real environment. Unfortunately, \emph{generalization tests in the real environment were not provided}. \emph{The instability and high variance} were addressed by subtracting a bias in the PG equation. The bias was defined as the average of job slowdowns taken at a single time step across all episodes. \emph{The implementation of DeepRM was open sourced allowing others to reproduce the results}. \emph{The rewards, actions, and states defined allowed the agent to learn a policy that performed well in the simulated environment}. Note that defining the state of the system was easier because the environment was simulated.  The solution also considered multiple reward definitions. For example, $-|J|$, where $J$ is the number of unfinished jobs in
the system. This reward definition optimizes the average job completion time. The jobs evaluated in DeepRM were considered to arrive online according to a Bernoulli process. In addition, the jobs were chosen randomly and it is unclear whether they represent real workload scenarios or not. This emphasizes the \emph{need for standardized benchmarks and frameworks} to evaluate the effectiveness of deep RL methods in scheduling jobs in the cloud.   
 \vspace{-0.25cm}
\section{Future Directions}
\vspace{-0.25cm}

We see multiple future directions for the deployment of deep RL in system optimization tasks. The general assumption is that deep RL may be useful in every system problem where the problem can be formulated as a sequential decision making process, and where meaningful action, state, and reward definitions can be provided. The objective of deep RL in such systems may span a wide range of options, such as energy efficiency, power, reliability, monitoring, revenue, performance, and utilization. At the processor level, deep RL could be used in branch prediction, memory prefetching, caching, data alignment, garbage collection, thread/task scheduling, power management, reliability, and monitoring. Compilers may also benefit from using deep RL to optimize the order of passes (optimizations), knobs/pragmas, unrolling factors, memory expansion, function inlining, vectorizing multiple instructions, tiling and instruction selection. With advancement of in- and near-memory processing, deep RL can be used to determine which portions of a workload should be performed in/near memory and which outside the memory.

At a higher system level, deep RL may be used in SQL/pandas query optimization, cloud computing, scheduling, caching, monitoring (e.g., temperature/failure) and fault tolerance, packet routing and classification, congestion control, FPGA allocation, and algorithm selection. While some of this has already been done, we believe there is big potential for improvement. It is necessary to explore more benchmarks, stable and generalizable learners, transfer learning approaches, RL algorithms, model-based RL and, more importantly, to provide better encoding of the states, actions and rewards to better represent the system and thus improve the learning. For example, with SQL/pandas join order optimization, the contents of the database are critical for determining the best order, and thus somehow incorporating an encoding of these contents may further improve the performance.

There is room for improvement in the RL algorithms as well. Some action and state spaces can dynamically change with time. For example, when adding a new node to a cluster, the RL agent will always skip the added node and it will not be captured in the environment state. Generally, the state transition function of the environment is unknown to the agent. Therefore, there is no guarantee that if the agent takes a certain action, a certain state will follow in the environment. This issue was presented in~\cite{kulkarni2012mitigating}, where compiler optimization passes were selected using deep RL. The authors mentioned a situation where the agent is stuck in an infinite loop of repeatedly picking the same optimization (action) back to back. This issue arose when a particular optimization did not change the features that describe the state of the environment, causing the neural network to apply the same optimization. To break this infinite loop, the authors limited the number of repetitions to five, and then instead, applied the second best optimization. This was done by taking the actions that corresponds to the second highest probability from the neural network's probability distribution output.

 \section{Conclusion} 
\label{conc}
\vspace{-0.25cm}

In this work, we reviewed and discussed multiple challenges in applying deep reinforcement learning to system optimization problems and proposed a set of metrics that can help evaluate the effectiveness of these solutions. Recent applications of deep RL in system optimization are mainly in packet classification, congestion control, compiler optimization, scheduling, query optimization and cloud computing. The growing complexity in systems demands learning based approaches. Deep RL presents unique opportunity to address the dynamic behavior of systems. Applying deep RL to systems proposes new set of challenges on how to frame and evaluate deep RL techniques. We anticipate that solving these challenges will enable system optimization with deep RL to grow.
%Looking forward, we anticipate that its application will grow. We also believe there is much more room for improvement in both the deep reinforcement learning algorithms and the system problems that could be tackled with it. 

%To that end, we plan to thoroughly tackle different challenges in compiler optimization and provide a comprehensive evaluation in a future work. 

% The \sysmlEqualContribution command is standard text for equal contribution.

 % leave blank if no need to mention equal contribution
%\printAffiliationsAndNotice{\sysmlEqualContribution} % otherwise use the standard text.

\bibliography{main}
\bibliographystyle{sysml2019}

\end{document}